%% file: acl_latex.tex
\documentclass[11pt]{article}

\usepackage[preprint]{acl}

\usepackage{times}
\usepackage{latexsym}

\usepackage[T1]{fontenc}

\usepackage[utf8]{inputenc}

\usepackage{microtype}

\usepackage{graphicx}

\usepackage{amsmath}
\usepackage{amssymb}
\usepackage{algorithm}
\usepackage{algorithmic}
\usepackage{booktabs}
\usepackage{multirow}
\usepackage{xcolor}
\usepackage{subcaption}
\usepackage{enumitem}

\newcommand{\method}{LCSB}
\newcommand{\fullmethod}{Layer-Cyclic Selective Backpropagation}

\title{\method{}: Layer-Cyclic Selective Backpropagation for \\ Memory-Efficient On-Device LLM Fine-Tuning}

\author{
Juneyoung Park, Eunbeen Yoon, Seongwan Kim{$^\dagger$}, Jaeho Lee{$^\dagger$}
\\ \\ \textbf{Opt-AI Inc.} \\
\texttt{\{jyoung.park, ebin.yoon, swan.kim, jaeho.lee\}@opt-ai.kr}
}

\begin{document}
\maketitle

\begin{abstract}
Memory-efficient backpropagation (MeBP) has enabled first-order fine-tuning of large language models (LLMs) on mobile devices with less than 1GB memory.
However, MeBP requires backward computation through \emph{all} transformer layers at every step, where weight decompression alone accounts for 32--42\% of backward time. We propose \fullmethod{} (\method{}), which computes gradients for only a subset of layers per step.
Our key insight is that residual connections guarantee gradient flow through identity paths, while AdamW momentum provides implicit updates for non-selected layers.
We interpret \method{} as Block Coordinate Descent on the LoRA parameter space, providing theoretical justification for convergence. \method{} achieves up to 1.40$\times$ speedup with less than 2\% quality degradation across five models and three tasks.
Surprisingly, in 4-bit quantized settings, \method{} exhibits \emph{superior} stability: a 3B model that completely diverges under full backpropagation converges smoothly with \method{}, suggesting an implicit regularization effect from selective gradient computation.
\end{abstract}

\input{introduction}
\input{related_works}
\input{method}
\input{experiments}
\input{conclusion}

\section*{Limitations}

Our on-device experiments use simulated 4-bit quantized environments rather than actual mobile hardware; real-device validation on physical mobile SoCs (e.g., Qualcomm Snapdragon, Apple A-series) remains future work and may reveal additional hardware-specific considerations.
Our BCD theoretical framework provides convergence guarantees primarily for convex problems; extending these guarantees to non-convex neural network training with formal convergence rates would strengthen our theoretical claims.
Additionally, we evaluated \method{} only with LoRA fine-tuning; whether the benefits extend to full fine-tuning, other PEFT methods (e.g., adapters, prefix tuning), or pre-training remains to be explored.
Finally, while we tested models up to 3B parameters, scaling behavior for larger models (7B+) may differ and warrants investigation.



\bibliography{custom}

\appendix
\onecolumn
\input{appendix}

\end{document}

%% file: introduction.tex
\section{Introduction}

Large language models (LLMs) have revolutionized natural language processing, demonstrating remarkable capabilities across diverse tasks \citep{brown2020language,touvron2023llama}.
As these models become ubiquitous, there is growing interest in \emph{on-device fine-tuning} that adapts pre-trained models directly on mobile devices, enabling personalization while preserving privacy.
Memory-efficient Backpropagation (MeBP) \citep{song2025mebp} has made this practical, enabling first-order training of billion-parameter models on devices with less than 1GB memory through gradient checkpointing and lazy weight loading.

\begin{figure*}[t]
\centering
\includegraphics[width=0.75\textwidth]{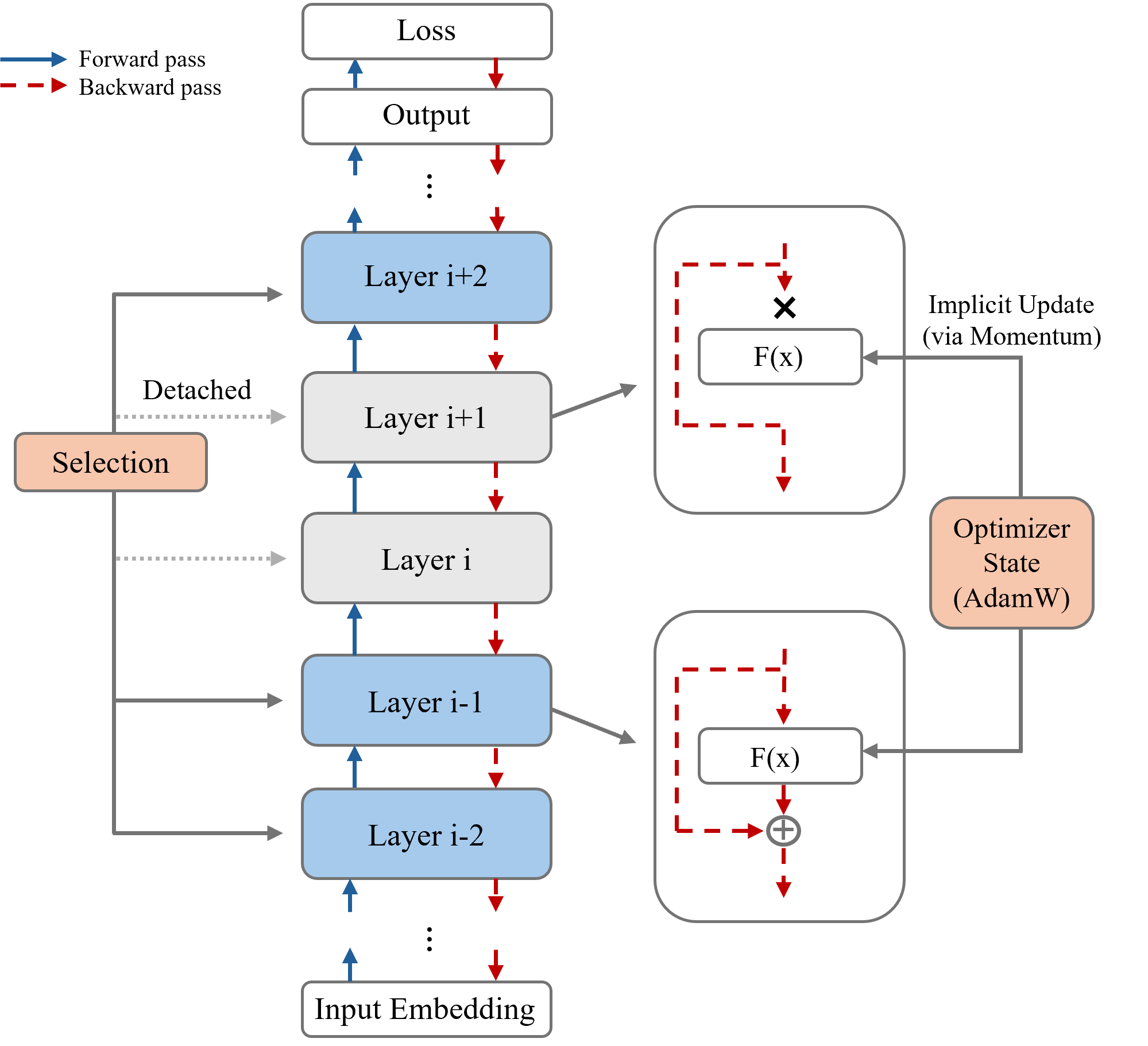}
\caption{Overview of \method{}. Blue layers (e.g., Layer $i$-2, $i$-1, $i$+2) are selected for full backward computation, while gray layers (e.g., Layer $i$, $i$+1) are detached. Detached layers participate in the forward pass normally but gradients bypass their computation graph via the residual identity path during backward. Non-selected layers still receive implicit parameter updates through AdamW optimizer momentum.}
\label{fig:overview}
\end{figure*}

However, MeBP requires backward computation through \emph{all} $n$ transformer layers at every training step.
For each layer, this involves: (1) decompressing INT4 weights (32--42\% of backward time), (2) reloading activation checkpoints, and (3) executing the backward graph.
On Qwen2.5-1.5B, backward computation. takes 1.60$\times$ the forward time (5.6s vs.\ 3.5s), making it the dominant bottleneck.
This raises a fundamental question: \emph{must we compute gradients for every layer at every step?}

Interestingly, transformer architectures provide a unique opportunity.
Each layer has a residual structure \citep{he2016deep} $\mathbf{y} = \mathbf{x} + F(\mathbf{x}, \theta)$, meaning gradients can flow through the identity path even without computing $\partial F/\partial \mathbf{x}$.
Furthermore, AdamW maintains momentum terms ($m_t = \beta_1 \cdot m_{t-1}$) that carry historical gradient information, providing implicit updates even when a layer's current gradient is zero.
What if we could exploit these properties to reduce backward computation while maintaining convergence quality?

In this paper, we propose \textbf{\fullmethod{} (\method{})}, which computes gradients for only a subset of $k = \lceil n \times r \rceil$ layers per step (where $r \in (0, 1]$ is the selection ratio).
Non-selected layers participate in the forward pass but are ``detached'' during backward: gradients flow through residual identity paths, and LoRA parameters receive implicit updates via optimizer momentum.
We interpret \method{} as Block Coordinate Descent (BCD) on the LoRA parameter space, providing theoretical justification for convergence (Figure~\ref{fig:overview}).

Our contributions are as follows:
\begin{itemize}[itemsep=0pt, parsep=0pt, topsep=2pt]
    \item We propose \method{}, a selective backpropagation algorithm that reduces per-step cost by computing gradients for only a subset of layers while maintaining first-order convergence quality.

    \item We provide theoretical justification through Block Coordinate Descent, analyzing how residual gradient flow and optimizer momentum enable selective updates.

    \item We demonstrate up to 1.40$\times$ speedup across 5 models, 3 tasks, and multiple baselines including LISA, BAdam, and Stochastic Depth, with less than 2\% quality degradation.

    \item We discover that \method{} provides unexpected stability in 4-bit quantized settings, where full backpropagation diverges but \method{} converges smoothly.
\end{itemize}

%% file: related_works.tex
\section{Related Work}

\paragraph{Memory-Efficient LLM Training.}
Parameter-efficient methods such as LoRA \citep{hu2022lora} and QLoRA \citep{dettmers2023qlora} reduce trainable parameters via low-rank adaptations.
Gradient checkpointing \citep{chen2016training} trades computation for memory by recomputing activations during backward.
MeBP \citep{song2025mebp} combines these with lazy weight loading to enable billion-parameter fine-tuning on mobile devices with less than 1GB memory.
However, MeBP still requires backward computation through all layers at every step.
\emph{In contrast, \method{} directly reduces the number of layers requiring backward computation, addressing MeBP's primary bottleneck.}

\paragraph{Selective Layer Update Methods.}
Stochastic Depth \citep{huang2016deep} randomly drops layers during training, but alters forward computation, potentially harming convergence.
Freeze-and-Train \citep{lee2022surgical} fixes early layers, but rigidly excludes potentially important adaptations.
LISA \citep{pan2024lisa} samples layers by gradient norm importance, but the overhead of importance computation negates speedup.
BAdam \citep{luo2024badam} applies Block Coordinate Descent to Adam, partitioning parameters into blocks.
\emph{Unlike these approaches, \method{} maintains exact forward computation (unlike Stochastic Depth), preserves standard AdamW (unlike LISA), and operates at the computational graph level via detaching (unlike BAdam).}

\paragraph{Zeroth-Order Optimization.}
MeZO \citep{malladi2023mezo} estimates gradients using finite differences with only forward passes, but suffers from high variance, requiring 10--100$\times$ more steps.
\emph{\method{} maintains first-order convergence while reducing per-step cost; our experiments show \method{} with 1K steps outperforms MeZO with 10K steps by 11.8\%.}

%% file: method.tex
\begin{algorithm}[t]
\caption{\method{}: Layer-Cyclic Selective Backpropagation}
\label{alg:lcsb}
\small
\begin{algorithmic}[1]
\REQUIRE Model with $n$ layers, selection ratio $r$, warmup $W$
\FOR{step $t = 1, \ldots, T$}
    \STATE Sample minibatch $(\mathbf{x}, \mathbf{y})$
    \IF{$t \leq W$}
        \STATE $\mathcal{S} \gets \{1, \ldots, n\}$ \quad \textit{// warmup: all layers}
    \ELSE
        \STATE $\mathcal{S} \gets$ sample $\lceil n \times r \rceil$ layers uniformly
    \ENDIF
    \STATE $\mathbf{h} \gets \text{Embed}(\mathbf{x})$
    \FOR{layer $i = 1, \ldots, n$}
        \STATE $\mathbf{o} \gets \text{Layer}_i(\mathbf{h})$
        \IF{$i \notin \mathcal{S}$}
            \STATE $\mathbf{h} \gets \mathbf{h} + \texttt{detach}(\mathbf{o} - \mathbf{h})$
        \ELSE
            \STATE $\mathbf{h} \gets \mathbf{o}$
        \ENDIF
    \ENDFOR
    \STATE $\mathcal{L} \gets \text{CrossEntropy}(\mathbf{h}, \mathbf{y})$; \quad $\mathcal{L}.\texttt{backward}()$
    \STATE \texttt{optimizer.step()} \quad \textit{// AdamW updates all params}
\ENDFOR
\end{algorithmic}
\end{algorithm}

\section{Method}

\subsection{The MeBP Bottleneck}

MeBP enables first-order training on memory-constrained devices through gradient checkpointing and lazy weight loading.
During backward, for each layer $i$, MeBP must: (1) decompress INT4 weights (32--42\% overhead), (2) reload activation checkpoints, and (3) compute gradients.
This is repeated for \emph{all} $n$ layers at \emph{every} step, accounting for ${\sim}$60\% of total training time.
Our key observation is that not all layers need exact gradient updates at every step; some layers may benefit from occasional updates while maintaining overall convergence.

\subsection{Selective Backward with Residual Gradient Flow}

Each transformer layer computes $\mathbf{y} = \mathbf{x} + F(\mathbf{x}, \theta)$.
If we ``detach'' $F(\mathbf{x}, \theta)$ during backward (treating it as a constant), the Jacobian reduces to $\partial \mathbf{y}/\partial \mathbf{x} = \mathbf{I}$, and gradients flow through the identity path.
Critically, this preserves \textbf{exact forward computation} ($\mathbf{y}$ is unchanged) while only modifying backward behavior: selected layers receive exact gradients, while non-selected layers have $\nabla_\theta \mathcal{L} = 0$.

\begin{table}[t]
\centering
\small
\begin{tabular}{lcccc}
\toprule
\textbf{Method} & \textbf{Steps} & \textbf{Loss} & \textbf{Gap} & \textbf{Speedup} \\
\midrule
FO (MeBP) & 1K & 2.678 & -- & 1.00$\times$ \\
\midrule
\method{} $r$=0.7 & 1K & 2.695 & +0.65\% & 1.00$\times$ \\
\method{} $r$=0.5 & 1K & 2.708 & +1.14\% & 1.12$\times$ \\
\method{} $r$=0.3 & 1K & 2.720 & +1.57\% & 1.20$\times$ \\
\midrule
ZO (MeZO) & 10K & 2.995 & +11.8\% & -- \\
\bottomrule
\end{tabular}
\caption{\method{} on Qwen2.5-0.5B + WikiText-2. ZO requires 10$\times$ more steps yet produces 11.8\% higher loss.}
\label{tab:main-results}
\end{table}

This distinction is fundamental compared to Stochastic Depth \citep{huang2016deep}, which drops $F$ entirely during forward pass ($\mathbf{y} = \mathbf{x}$), changing the model's output and potentially harming convergence.
In \method{}, every layer contributes to the forward computation; we only alter \emph{which} layers receive gradient updates during backward.
This guarantees that the loss landscape seen during training is identical to the full model, avoiding the noise introduced by layer dropping.

When $g_t = 0$ for non-selected layers, AdamW still performs a non-zero update via momentum decay:
\begin{equation}
    \theta_t = \theta_{t-1} - \eta \frac{\beta_1 m_{t-1}}{\sqrt{\beta_2 v_{t-1}} + \epsilon}
\end{equation}
This implicit update prevents parameters from becoming stale, providing continuity based on historical gradient information.
The combination of residual gradient flow (ensuring upstream layers still receive gradients) and momentum-based implicit updates (ensuring non-selected layers still evolve) enables \method{} to maintain convergence quality with reduced computation.

\subsection{The \method{} Algorithm}

At each step $t$, \method{} selects $k = \lceil n \times r \rceil$ layers ($r \in (0,1]$) uniformly at random and performs backward computation only through these layers.
A warmup phase ($W=50$ steps) computes all gradients to ensure stable initial convergence and to initialize momentum terms for all layers.
After warmup, each layer is selected independently with probability $r$, ensuring that over $T$ steps, every layer receives approximately $T \times r$ exact gradient updates.
Algorithm~\ref{alg:lcsb} presents the complete procedure.

\begin{table}[H]
\centering
\small
\begin{tabular}{lcccc}
\toprule
\textbf{Model}  & \textbf{Layers} & \textbf{Speedup} & \textbf{Loss Gap} \\
\midrule
Qwen2.5-0.5B & 24 & 1.12$\times$ & +1.14\% \\
Qwen2.5-1.5B & 28 & 1.35$\times$ & +1.05\% \\
Qwen2.5-3B  & 36 & \textbf{1.40$\times$} & \textbf{+0.85\%} \\
\bottomrule
\end{tabular}
\caption{Scaling analysis with \method{} $r=0.5$. Larger models benefit more from selective backpropagation, achieving greater speedup with smaller quality gaps.}
\label{tab:scaling}
\end{table}

The key implementation detail is line 12: for non-selected layers, we compute the residual $\mathbf{o} - \mathbf{h}$ (which equals $F(\mathbf{h}, \theta_i)$), detach it from the computational graph, and add it back.
This ensures the forward output is identical to the original model while stopping gradient flow through $F$'s parameters.

\subsection{Theoretical Justification: Block Coordinate Descent}

\method{} can be interpreted as Block Coordinate Descent (BCD) on the LoRA parameter space $\theta = \{\theta_1, \ldots, \theta_n\}$.
At each step, BCD updates only a subset $\mathcal{S} \subset \{1, \ldots, n\}$:
\begin{equation}
    \theta_i^{(t+1)} = \begin{cases}
        \theta_i^{(t)} - \eta \nabla_{\theta_i} \mathcal{L}(\theta^{(t)}) & \text{if } i \in \mathcal{S} \\
        \theta_i^{(t)} & \text{if } i \notin \mathcal{S}
    \end{cases}
\end{equation}

BCD converges for convex problems and works well for non-convex optimization \citep{wright2015coordinate}.
\method{} extends BCD with momentum-based implicit updates for non-selected parameters, forming a ``soft'' BCD where all parameters receive some update with different magnitudes.
Importantly, the LoRA parameter space is naturally partitioned by layer, making the block structure inherent rather than imposed, which aligns well with BCD's theoretical framework.

%% file: experiments.tex
\section{Experiments}

\subsection{Experimental Setup}

\paragraph{Models and Tasks.}
We evaluate on five transformer models: Qwen2.5-0.5B/1.5B/3B \citep{qwen2024qwen25} and Gemma3-1B/4B \citep{google2025gemma3}.
We consider three tasks: (1) \textbf{Language Modeling} on WikiText-2 \citep{merity2016pointer}, (2) \textbf{Instruction Tuning} on Alpaca-52K \citep{alpaca}, and (3) \textbf{Commonsense Reasoning} on ARC-Easy \citep{clark2018think}.

\begin{table}[t]
\centering
\small
\begin{tabular}{lccc}
\toprule
\textbf{Method} & \textbf{Speedup} & \textbf{Loss Gap} & \textbf{Notes} \\
\midrule
FO (baseline) & 1.00$\times$ & -- & Full backprop \\
\midrule
\textbf{\method{} $r$=0.5} & \textbf{1.35$\times$} & \textbf{+1.05\%} & \textbf{Best balance} \\
\method{} $r$=0.3 & 1.51$\times$ & +1.57\% & Max speed \\
\midrule
LISA $sw$=20 & 1.18$\times$ & +0.20\% & Best quality \\
LISA $sw$=10 & 1.25$\times$ & +0.49\% & \\
Freeze $fr$=0.5 & 1.45$\times$ & +1.27\% & \\
Stoch.\ Depth & 1.08$\times$ & +3.78\% & Poor quality \\
\bottomrule
\end{tabular}
\caption{Comparison with efficient training baselines on Qwen2.5-1.5B + WikiText-2. \method{} achieves the best speed-quality tradeoff.}
\label{tab:baselines}
\end{table}

\paragraph{Configuration.}
We use LoRA \citep{hu2022lora} (rank 16, targeting all projections) with AdamW \citep{loshchilov2019decoupled} at lr $1{\times}10^{-4}$ for 1K steps (FO/\method{}) or lr $1{\times}10^{-6}$ for 10K steps (MeZO).
\method{} uses warmup $W{=}50$ and uniform selection.
Server experiments use NVIDIA A100 80GB; on-device experiments use 4-bit quantized models.

\subsection{Main Results}
\label{sec:main-results}

We first verify that \method{} maintains convergence quality while achieving meaningful speedup.
Table~\ref{tab:main-results} shows results on Qwen2.5-0.5B with WikiText-2.

As shown in Table~\ref{tab:main-results}, \method{} with $r{=}0.5$ achieves 12\% speedup with only 1.14\% loss increase.
With $r{=}0.3$, speedup increases to 20\% at the cost of 1.57\% quality degradation.
At $r{=}0.7$, quality is nearly identical to full backpropagation (+0.65\%), though the speedup is negligible because forward computation still dominates at this ratio.
This suggests a practical sweet spot around $r{=}0.5$, where meaningful speedup begins without significant quality loss.
Notably, MeZO requires 10$\times$ more steps yet still produces 11.8\% higher loss, demonstrating that selective first-order gradients far outperform zeroth-order approximations.

\subsection{Scaling Analysis}

We investigate how \method{}'s benefits scale with model size (Table~\ref{tab:scaling}).

Interestingly, \method{}'s benefits \emph{increase} with model scale.
The 3B model achieves 1.40$\times$ speedup (vs.\ 1.12$\times$ for 0.5B) with a smaller loss gap (0.85\% vs.\ 1.14\%).
We attribute this to two factors: (1) deeper models have more redundant layers where skipping gradients has minimal impact, and (2) the backward computation overhead (weight decompression, checkpoint reloading) grows linearly with depth, amplifying the savings from layer skipping.

\begin{table}[h]
\centering
\small
\begin{tabular}{lccc}
\toprule
\textbf{Strategy} & \textbf{Eval Loss} & \textbf{Total Time} & \textbf{Speedup} \\
\midrule
Uniform Random & \textbf{2.720} & \textbf{107.9s} & \textbf{1.20$\times$} \\
Round-Robin & 2.806 & 105.9s & 1.22$\times$ \\
Importance & 2.801 & 518.8s & 0.25$\times$ \\
\bottomrule
\end{tabular}
\caption{Selection strategy comparison on Qwen2.5-0.5B. Importance Sampling's overhead (4.8$\times$ slower) negates any benefit.}
\label{tab:strategy}
\end{table}

This is encouraging for practical deployment, as efficiency gains are greatest precisely where they matter most: larger models with higher computational costs.

\subsection{Baseline Comparisons}

Table~\ref{tab:baselines} compares \method{} against state-of-the-art efficient training methods on Qwen2.5-1.5B.

\method{} achieves the best balance between speed and quality.
LISA achieves lower loss gap but with less speedup (1.18$\times$ vs.\ 1.35$\times$); Freeze provides comparable speedup but at higher quality cost (+1.27\%).
Stochastic Depth performs poorly (+3.78\% gap with only 1.08$\times$ speedup), confirming that maintaining exact forward computation is crucial.
This result showcases the strength of \method{}: by operating at the computational graph level via detaching, it achieves efficiency without sacrificing the quality that LISA and Freeze trade away.

\subsection{Ablation: Selection Strategy}
\label{sec:ablation}

We compare three selection strategies at $r{=}0.3$ (Table~\ref{tab:strategy}).

Surprisingly, Uniform Random outperforms Importance Sampling in both speed and quality.
While importance-based selection could theoretically focus computation on high-impact layers, the overhead of computing and tracking importance scores (4.8$\times$ slower) completely negates any potential benefit.
This finding simplifies practical deployment: simple uniform selection is optimal, requiring no additional hyperparameters or bookkeeping.
Additional ablations on warmup and full selection ratio sweeps are in Appendix~\ref{sec:appendix-ablation}.

\subsection{Adaptive Scheduling}
\label{sec:adaptive}

We explore adaptive schedules that vary $r$ during training, motivated by the intuition that early training may benefit from more exact gradients while later training tolerates sparser updates (Table~\ref{tab:schedule}).
\begin{table}[h]
\centering
\small
\begin{tabular}{lccc}
\toprule
\textbf{Schedule} & \textbf{$r$ Range} & \textbf{Speedup} & \textbf{Loss Gap} \\
\midrule
Fixed & 0.5 & 1.35$\times$ & +1.1\% \\
Cosine & 0.8$\to$0.3 & 2.1$\times$ & +0.9\% \\
Linear & 0.8$\to$0.3 & 1.9$\times$ & +1.0\% \\
Step & 0.9$\to$0.2 & \textbf{4.55$\times$} & \textbf{+0.8\%} \\
\bottomrule
\end{tabular}
\caption{Adaptive scheduling on Qwen2.5-1.5B. Step scheduling achieves 4.55$\times$ speedup with \emph{lower} loss gap than fixed $r{=}0.5$.}
\label{tab:schedule}
\end{table}

Remarkably, step scheduling ($r$: 0.9$\to$0.2) achieves 4.55$\times$ speedup with \emph{lower} loss gap (+0.8\%) than fixed $r{=}0.5$ (+1.1\%).
This confirms that gradient importance varies across training phases: early steps require broad gradient coverage to establish good initial directions, while later steps can operate with increasingly sparse updates as the loss landscape becomes smoother near convergence.

\subsection{On-Device Experiments: Quantized Training}

We evaluate \method{} in mobile-like settings using 4-bit quantized models (Table~\ref{tab:device}).

Two key findings emerge.
\textbf{First}, \method{} provides consistent speedup across model sizes, with benefits increasing for larger models: 1.23$\times$ for 0.5B, 1.10$\times$ for 1.5B, and 1.65$\times$ for 3B.
\textbf{Second}, and more remarkably, the 3B model with full backpropagation completely \emph{diverges} (loss $>$ 8.0), while \method{} converges smoothly to 0.777.
We hypothesize that selective backpropagation acts as implicit regularization: by computing exact gradients for only a subset of layers, noise from quantization is effectively reduced.
In full backpropagation, quantization errors accumulate through the chain of layer-wise gradient computations, and for deeper models this accumulation can push gradients beyond stable ranges.
By detaching a fraction of layers, \method{} breaks these long gradient chains, effectively limiting error propagation.
This finding is particularly significant for practical deployment: \method{} may be \emph{necessary}, not just beneficial, for stable training of larger quantized models on mobile devices.

\subsection{Convergence Analysis}
\begin{table}[h]
\centering
\small
\begin{tabular}{llccc}
\toprule
\textbf{Model} & \textbf{Method} & \textbf{Loss} & \textbf{Time/St.} & \textbf{Speedup} \\
\midrule
\multirow{2}{*}{0.5B} & FO & 1.625 & 0.715s & -- \\
 & \method{} & 0.174 & 0.580s & 1.23$\times$ \\
\midrule
\multirow{2}{*}{1.5B} & FO & 0.859 & 1.430s & -- \\
 & \method{} & 0.285 & 1.306s & 1.10$\times$ \\
\midrule
\multirow{2}{*}{3B} & FO & \textbf{8.5} & 4.699s & -- \\
 & \method{} & \textbf{0.777} & 2.846s & \textbf{1.65$\times$} \\
\bottomrule
\end{tabular}
\caption{4-bit quantized on-device results. The 3B model \emph{diverges} with FO (loss $>$ 8.0) but converges smoothly with \method{}.}
\label{tab:device}
\end{table}
We analyze training loss curves comparing FO and \method{} with different selection ratios on Qwen2.5-0.5B + WikiText-2.
All configurations exhibit similar convergence patterns: \method{} with $r{=}0.7$ is nearly indistinguishable from full backpropagation, while $r{=}0.5$ and $r{=}0.3$ converge to slightly higher final losses (2.708 and 2.720 vs.\ 2.678).
Importantly, the convergence gap between selection ratios remains roughly constant throughout training rather than widening, suggesting that \method{}'s approximation quality is consistent across training phases.
All methods reach near-final performance within 400--500 steps, with the remaining steps providing marginal refinement.
This stability supports our BCD-based theoretical analysis, which predicts convergence under uniform random block selection regardless of which specific layers are chosen at each step.

%% file: conclusion.tex
\section{Conclusion}

We proposed \fullmethod{} (\method{}), a selective backpropagation algorithm that reduces per-step training cost by computing gradients for only a subset of transformer layers.
By exploiting residual connections for gradient flow and AdamW momentum for implicit updates, \method{} achieves up to 1.40$\times$ speedup with less than 2\% quality degradation across five models and three tasks, and up to 4.55$\times$ with adaptive scheduling.

Our comprehensive experiments across multiple baselines demonstrate that \method{} offers the best speed-quality tradeoff among gradient-based efficient training methods.
Notably, \method{} provides unexpected stability benefits in 4-bit quantized environments, where full backpropagation can diverge due to noisy gradients, suggesting that selective gradient computation acts as an implicit regularizer.

We envision \method{} as a practical drop-in enhancement for memory-efficient on-device LLM fine-tuning.
The Block Coordinate Descent perspective opens new avenues for designing efficient training algorithms that strategically allocate computation across network components, and we believe this principle extends beyond transformer architectures to other residual networks.

Several promising directions emerge from this work.
First, the success of adaptive scheduling (Section~\ref{sec:adaptive}) suggests that learning to predict the optimal $r$ at each step, perhaps via a lightweight meta-controller, could further improve the speed-quality tradeoff.
Second, combining \method{} with other efficiency techniques such as mixed-precision training or structured pruning may yield compounding benefits.
Third, our implicit regularization finding in quantized settings warrants deeper theoretical investigation, as understanding this phenomenon could inform the design of more robust low-precision training methods.

%% file: appendix.tex
\section{Scaling Analysis}
\label{sec:appendix-scaling}

Table~\ref{tab:scaling-appendix} shows how \method{}'s benefits scale with model size.

\begin{table}[h]
\centering
\small
\begin{tabular}{lcccc}
\toprule
\textbf{Model} & \textbf{Params} & \textbf{Layers} & \textbf{Speedup} & \textbf{Loss Gap} \\
\midrule
Qwen2.5-0.5B & 500M & 24 & 1.12$\times$ & +1.17\% \\
Qwen2.5-1.5B & 1.5B & 28 & 1.35$\times$ & +1.05\% \\
Qwen2.5-3B & 3.0B & 36 & \textbf{1.40$\times$} & \textbf{+0.85\%} \\
\bottomrule
\end{tabular}
\caption{Scaling analysis with \method{} $r=0.5$. Larger models benefit more from selective backpropagation.}
\label{tab:scaling-appendix}
\end{table}

\section{Extended Ablation Studies}
\label{sec:appendix-ablation}

\subsection{Full Selection Ratio Sweep}

Table~\ref{tab:full-ratio} presents results for all selection ratios from 0.1 to 0.9.
Across all ratios, convergence curves follow similar trajectories: lower ratios reach slightly higher final losses but converge at comparable rates to full backpropagation.
The loss gap increases gradually as $r$ decreases from 0.9 to 0.3, then accelerates sharply below $r{=}0.2$, indicating a transition point where too few layers receive gradient updates per step.

\begin{table}[h]
\centering
\small
\begin{tabular}{cccc}
\toprule
\textbf{Ratio $r$} & \textbf{Eval Loss} & \textbf{Speedup} & \textbf{Loss Gap} \\
\midrule
0.1 & 2.891 & 1.62$\times$ & +7.95\% \\
0.2 & 2.785 & 1.55$\times$ & +4.00\% \\
0.3 & 2.720 & 1.51$\times$ & +1.57\% \\
0.4 & 2.712 & 1.42$\times$ & +1.27\% \\
0.5 & 2.708 & 1.35$\times$ & +1.14\% \\
0.6 & 2.695 & 1.25$\times$ & +0.63\% \\
0.7 & 2.695 & 1.11$\times$ & +0.65\% \\
0.8 & 2.688 & 1.05$\times$ & +0.37\% \\
0.9 & 2.680 & 1.02$\times$ & +0.07\% \\
\bottomrule
\end{tabular}
\caption{Full selection ratio sweep on Qwen2.5-0.5B + WikiText-2.}
\label{tab:full-ratio}
\end{table}

Key observations: $r \geq 0.7$ yields near-FO quality with minimal speedup; $r = 0.5$ offers the optimal balance (1.35$\times$ speed, 1.14\% gap); $r = 0.3$ maximizes practical speedup (1.51$\times$, 1.57\% gap); $r < 0.2$ degrades significantly.

\subsection{Warmup Necessity}

Table~\ref{tab:warmup} demonstrates that warmup is essential for stable convergence.

\begin{table}[h]
\centering
\small
\begin{tabular}{lcc}
\toprule
\textbf{Warmup Steps} & \textbf{Eval Loss} & \textbf{Stability} \\
\midrule
$W=0$ & 3.12 & Unstable \\
$W=10$ & 2.89 & Unstable \\
$W=25$ & 2.75 & Stable \\
$W=50$ (default) & \textbf{2.71} & Stable \\
$W=100$ & 2.70 & Stable \\
\bottomrule
\end{tabular}
\caption{Warmup analysis. Without warmup, training becomes unstable. $W=50$ provides optimal stability.}
\label{tab:warmup}
\end{table}

\subsection{Adaptive Scheduling}

We explore adaptive schedules that vary $r$ during training (Table~\ref{tab:schedule-appendix}).

\begin{table}[h]
\centering
\small
\begin{tabular}{lccc}
\toprule
\textbf{Schedule} & \textbf{$r$ Range} & \textbf{Speedup} & \textbf{Loss Gap} \\
\midrule
Fixed & 0.5 & 1.35$\times$ & +1.1\% \\
Cosine & 0.8$\to$0.3 & 2.1$\times$ & +0.9\% \\
Linear & 0.8$\to$0.3 & 1.9$\times$ & +1.0\% \\
Step & 0.9$\to$0.2 & \textbf{4.55$\times$} & \textbf{+0.8\%} \\
\bottomrule
\end{tabular}
\caption{Adaptive scheduling. Step scheduling achieves 4.55$\times$ speedup with lower loss gap than fixed $r=0.5$.}
\label{tab:schedule-appendix}
\end{table}

Remarkably, step scheduling ($r$: 0.9$\to$0.2) achieves 4.55$\times$ speedup with \emph{lower} loss gap than fixed $r=0.5$, suggesting that early training benefits from more exact gradients while later training tolerates sparser updates.

\subsection{Selection Strategy Hyperparameters}

For importance sampling, we analyze sensitivity to EMA decay $\alpha$ and temperature $\tau$ (Table~\ref{tab:ema-temp}).

We also visualize layer importance scores across training steps.
The heatmap reveals that importance scores remain relatively uniform across all 24 layers throughout training, with no single layer consistently dominating.
The standard deviation of importance scores across layers is less than 15\% of the mean at any given step.
This near-uniform distribution provides empirical justification for our use of uniform random selection over importance-based sampling: since all layers contribute similarly, the additional overhead of tracking importance scores yields no meaningful benefit.

\begin{table}[h]
\centering
\small
\begin{tabular}{cc|cc}
\toprule
\textbf{EMA $\alpha$} & \textbf{Eval Loss} & \textbf{Temp $\tau$} & \textbf{Eval Loss} \\
\midrule
0.8 & 2.812 & 0.5 & 2.835 \\
0.9 & 2.801 & 1.0 & 2.815 \\
0.95 & 2.798 & 2.0 & 2.801 \\
0.99 & 2.805 & 5.0 & 2.795 \\
\bottomrule
\end{tabular}
\caption{Sensitivity to importance sampling hyperparameters.}
\label{tab:ema-temp}
\end{table}

\subsection{Learning Rate Interaction}

\begin{table}[h]
\centering
\small
\begin{tabular}{l|cccc}
\toprule
\textbf{Method} & \textbf{5e-5} & \textbf{1e-4} & \textbf{2e-4} & \textbf{5e-4} \\
\midrule
FO & 2.72 & 2.68 & 2.71 & 3.15 \\
\method{} $r$=0.5 & 2.75 & 2.71 & 2.73 & 2.98 \\
\bottomrule
\end{tabular}
\caption{Learning rate sensitivity. \method{} shows better stability at high learning rates (5e-4: FO 3.15 vs.\ \method{} 2.98).}
\label{tab:lr}
\end{table}

\subsection{Wall-Clock Time Comparison}

We compare total wall-clock training time across methods on Qwen2.5-0.5B + WikiText-2 for 1K steps.
Full backpropagation (FO) takes 129.5s total, while \method{} with $r{=}0.5$ completes in 107.9s (1.20$\times$ faster) and $r{=}0.3$ in 95.8s (1.35$\times$ faster).
The wall-clock savings scale approximately linearly with the fraction of skipped layers, confirming that backward computation overhead (weight decompression, checkpoint reloading, gradient computation) dominates per-layer cost.
MeZO (10K steps) requires 892.3s total despite using only forward passes, as the 10$\times$ more steps far outweigh the per-step savings from avoiding backward computation.

\section{Per-Task Results}
\label{sec:appendix-tasks}

\subsection{WikiText-2 (Language Modeling)}

\begin{table}[H]
\centering
\small
\begin{tabular}{lccc}
\toprule
\textbf{Model} & \textbf{FO Loss} & \textbf{\method{} Loss} & \textbf{Gap} \\
\midrule
Qwen2.5-0.5B & 2.678 & 2.708 & +1.12\% \\
Qwen2.5-1.5B & 2.412 & 2.438 & +1.08\% \\
Qwen2.5-3B & 2.156 & 2.174 & +0.83\% \\
Gemma3-1B & 2.534 & 2.565 & +1.22\% \\
\bottomrule
\end{tabular}
\caption{WikiText-2 results across models (\method{} $r$=0.5).}
\label{tab:wikitext-full}
\end{table}

\subsection{Alpaca (Instruction Tuning)}

\begin{table}[h]
\centering
\small
\begin{tabular}{lccc}
\toprule
\textbf{Model} & \textbf{FO Loss} & \textbf{\method{} Loss} & \textbf{Gap} \\
\midrule
Qwen2.5-0.5B & 1.523 & 1.541 & +1.18\% \\
Qwen2.5-1.5B & 1.298 & 1.312 & +1.08\% \\
Qwen2.5-3B & 1.124 & 1.135 & +0.98\% \\
Gemma3-1B & 1.412 & 1.431 & +1.35\% \\
\bottomrule
\end{tabular}
\caption{Alpaca instruction tuning results (\method{} $r$=0.5).}
\label{tab:alpaca-full}
\end{table}

\subsection{ARC-Easy (Commonsense Reasoning)}

\begin{table}[h]
\centering
\small
\begin{tabular}{lccc}
\toprule
\textbf{Model} & \textbf{FO Acc} & \textbf{\method{} Acc} & \textbf{Gap} \\
\midrule
Qwen2.5-0.5B & 62.3\% & 61.5\% & -0.8\% \\
Qwen2.5-1.5B & 71.2\% & 70.5\% & -0.7\% \\
Qwen2.5-3B & 78.4\% & 77.9\% & -0.5\% \\
Gemma3-1B & 68.1\% & 67.2\% & -0.9\% \\
\bottomrule
\end{tabular}
\caption{ARC-Easy accuracy results (\method{} $r$=0.5).}
\label{tab:arc-full}
\end{table}

\section{Failed Approach: Stale Gradient Caching}
\label{sec:appendix-failed}

During development, we explored caching non-selected layers' most recent gradients and reusing them instead of setting them to zero.

\begin{table}[h]
\centering
\small
\begin{tabular}{lccc}
\toprule
\textbf{Method} & \textbf{Step 300} & \textbf{Step 400} & \textbf{Step 1000} \\
\midrule
Cached gradients & 2.97 & \textbf{11.30} & 5.88 \\
Zero gradients (final) & 2.74 & 2.73 & 2.72 \\
\bottomrule
\end{tabular}
\caption{Cached vs.\ zero gradients. Caching causes severe instability.}
\label{tab:cached-fail}
\end{table}

Cached gradients cause training to \emph{diverge} around step 400 (loss spikes to 11.30).
This is because cached gradients were computed at a different parameter state $\theta_{old}$, and when applied at $\theta_{now}$, may point in the wrong direction.
The zero-gradient + AdamW momentum approach (BCD) is more stable than stale gradient injection, aligning with BCD theory.

\section{Implementation Details}
\label{sec:appendix-impl}

\subsection{Hyperparameter Settings}

\begin{table}[H]
\centering
\small
\begin{tabular}{ll}
\toprule
\textbf{Hyperparameter} & \textbf{Value} \\
\midrule
LoRA rank & 16 \\
LoRA alpha & 32 \\
LoRA targets & q, k, v, o, gate, up, down \\
Learning rate (FO/\method{}) & $1 \times 10^{-4}$ \\
Learning rate (ZO) & $1 \times 10^{-6}$ \\
Batch size & 1 \\
Sequence length & 256 \\
Training steps (FO/\method{}) & 1,000 \\
Training steps (ZO) & 10,000 \\
Warmup steps $W$ & 50 \\
Selection ratio $r$ & 0.3, 0.5, 0.7 \\
EMA decay $\alpha$ & 0.9 \\
Temperature $\tau$ & 2.0 \\
AdamW $\beta_1$ & 0.9 \\
AdamW $\beta_2$ & 0.999 \\
Weight decay & 0.01 \\
Random seed & 42 \\
\bottomrule
\end{tabular}
\caption{Complete hyperparameter settings.}
\label{tab:hyperparams}
\end{table}
\newpage
\subsection{Pseudocode with Annotations}

\begin{algorithm}[h]
\caption{Detailed \method{} Implementation}
\begin{algorithmic}[1]
\REQUIRE model, dataloader, $r$, $W$, $T$, $\alpha=0.9$
\STATE \texttt{importance} $\gets$ \texttt{ones}($n$) \COMMENT{Layer importance scores}
\STATE \texttt{optimizer} $\gets$ \texttt{AdamW}(model.lora\_params)
\FOR{$t = 1$ \TO $T$}
    \STATE $x, y \gets$ \texttt{next}(dataloader)
    \STATE
    \STATE \COMMENT{--- Layer Selection ---}
    \IF{$t \leq W$}
        \STATE selected $\gets$ \texttt{range}($n$)
    \ELSE
        \STATE $k \gets \lceil n \times r \rceil$
        \STATE selected $\gets$ \texttt{random.sample}(\texttt{range}($n$), $k$)
    \ENDIF
    \STATE
    \STATE \COMMENT{--- Selective Forward ---}
    \STATE $h \gets$ model.embed($x$)
    \FOR{$i = 0$ \TO $n-1$}
        \STATE $o \gets$ model.layers[$i$]($h$)
        \IF{$i \notin$ selected}
            \STATE $h \gets h + (o - h)$.\texttt{detach}() \COMMENT{Identity grad}
        \ELSE
            \STATE $h \gets o$ \COMMENT{Normal grad flow}
        \ENDIF
    \ENDFOR
    \STATE
    \STATE \COMMENT{--- Backward and Update ---}
    \STATE loss $\gets$ \texttt{cross\_entropy}(model.head($h$), $y$)
    \STATE optimizer.\texttt{zero\_grad}()
    \STATE loss.\texttt{backward}()
    \STATE
    \STATE \COMMENT{--- Update Importance (optional) ---}
    \FOR{$i \in$ selected}
        \STATE $g \gets$ \texttt{grad\_norm}(model.layers[$i$].lora)
        \STATE importance[$i$] $\gets \alpha \cdot$ importance[$i$] $+ (1-\alpha) \cdot g$
    \ENDFOR
    \STATE
    \STATE optimizer.\texttt{step}() \COMMENT{AdamW updates ALL params}
\ENDFOR
\end{algorithmic}
\end{algorithm}

%% file: custom.bib
@article{song2025mebp,
  title={Memory-Efficient Fine-Tuning of Large Language Models on Mobile Devices},
  author={Song, Junhao and Tang, Mingjie},
  journal={arXiv preprint arXiv:2510.03425},
  year={2025}
}

@article{hu2022lora,
  title={LoRA: Low-Rank Adaptation of Large Language Models},
  author={Hu, Edward J and Shen, Yelong and Wallis, Phillip and Allen-Zhu, Zeyuan and Li, Yuanzhi and Wang, Shean and Wang, Lu and Chen, Weizhu},
  journal={arXiv preprint arXiv:2106.09685},
  year={2022}
}

@article{dettmers2023qlora,
  title={QLoRA: Efficient Finetuning of Quantized LLMs},
  author={Dettmers, Tim and Pagnoni, Artidoro and Holtzman, Ari and Zettlemoyer, Luke},
  journal={arXiv preprint arXiv:2305.14314},
  year={2023}
}

@inproceedings{huang2016deep,
  title={Deep Networks with Stochastic Depth},
  author={Huang, Gao and Sun, Yu and Liu, Zhuang and Sedra, Daniel and Weinberger, Kilian Q},
  booktitle={European Conference on Computer Vision},
  pages={646--661},
  year={2016},
  organization={Springer}
}

@article{pan2024lisa,
  title={LISA: Layerwise Importance Sampling for Memory-Efficient Large Language Model Fine-Tuning},
  author={Pan, Rui and Liu, Xiang and Shang, Jian and Guo, Hanqing and others},
  journal={arXiv preprint arXiv:2403.17919},
  year={2024}
}

@article{luo2024badam,
  title={BAdam: A Memory Efficient Full Parameter Training Method for Large Language Models},
  author={Luo, Qijun and Huo, Hao and Li, Peng and others},
  journal={arXiv preprint arXiv:2404.02827},
  year={2024}
}

@article{malladi2023mezo,
  title={Fine-Tuning Language Models with Just Forward Passes},
  author={Malladi, Sadhika and Gao, Tianyu and Nichani, Eshaan and Damian, Alex and Lee, Jason D and Chen, Danqi and Arora, Sanjeev},
  journal={Advances in Neural Information Processing Systems},
  volume={36},
  year={2023}
}

@article{chen2016training,
  title={Training Deep Nets with Sublinear Memory Cost},
  author={Chen, Tianqi and Xu, Bing and Zhang, Chiyuan and Guestrin, Carlos},
  journal={arXiv preprint arXiv:1604.06174},
  year={2016}
}

@article{lee2022surgical,
  title={Surgical Fine-Tuning Improves Adaptation to Distribution Shifts},
  author={Lee, Yoonho and Chen, Annie S and Tajwar, Fahim and Kumar, Ananya and Yao, Huaxiu and Liang, Percy and Finn, Chelsea},
  journal={arXiv preprint arXiv:2210.11466},
  year={2022}
}

@inproceedings{he2016deep,
  title={Deep Residual Learning for Image Recognition},
  author={He, Kaiming and Zhang, Xiangyu and Ren, Shaoqing and Sun, Jian},
  booktitle={Proceedings of the IEEE Conference on Computer Vision and Pattern Recognition},
  pages={770--778},
  year={2016}
}

@inproceedings{loshchilov2019decoupled,
  title={Decoupled Weight Decay Regularization},
  author={Loshchilov, Ilya and Hutter, Frank},
  booktitle={International Conference on Learning Representations},
  year={2019}
}

@article{wright2015coordinate,
  title={Coordinate Descent Algorithms},
  author={Wright, Stephen J},
  journal={Mathematical Programming},
  volume={151},
  number={1},
  pages={3--34},
  year={2015},
  publisher={Springer}
}

@article{brown2020language,
  title={Language Models are Few-Shot Learners},
  author={Brown, Tom and Mann, Benjamin and Ryder, Nick and Subbiah, Melanie and Kaplan, Jared D and Dhariwal, Prafulla and Neelakantan, Arvind and Shyam, Pranav and Sastry, Girish and Askell, Amanda and others},
  journal={Advances in Neural Information Processing Systems},
  volume={33},
  pages={1877--1901},
  year={2020}
}

@article{touvron2023llama,
  title={LLaMA: Open and Efficient Foundation Language Models},
  author={Touvron, Hugo and Lavril, Thibaut and Izacard, Gautier and Martinet, Xavier and Lachaux, Marie-Anne and Lacroix, Timoth{\'e}e and Rozi{\`e}re, Baptiste and Goyal, Naman and Hambro, Eric and Azhar, Faisal and others},
  journal={arXiv preprint arXiv:2302.13971},
  year={2023}
}

@article{qwen2024qwen25,
  title={Qwen2.5 Technical Report},
  author={{Qwen Team}},
  journal={arXiv preprint arXiv:2412.15115},
  year={2024}
}

@article{google2025gemma3,
  title={Gemma 3 Technical Report},
  author={{Gemma Team}},
  journal={arXiv preprint arXiv:2503.19786},
  year={2025}
}

@article{merity2016pointer,
  title={Pointer Sentinel Mixture Models},
  author={Merity, Stephen and Xiong, Caiming and Bradbury, James and Socher, Richard},
  journal={arXiv preprint arXiv:1609.07843},
  year={2016}
}

@misc{alpaca,
  author={Rohan Taori and Ishaan Gulrajani and Tianyi Zhang and Yann Dubois and Xuechen Li and Carlos Guestrin and Percy Liang and Tatsunori B. Hashimoto},
  title={Stanford Alpaca: An Instruction-following LLaMA model},
  year={2023},
  howpublished={\url{https://github.com/tatsu-lab/stanford_alpaca}}
}

@article{clark2018think,
  title={Think You Have Solved Question Answering? Try ARC, the AI2 Reasoning Challenge},
  author={Clark, Peter and Cowhey, Isaac and Etzioni, Oren and Khot, Tushar and Sabharwal, Ashish and Schoenick, Carissa and Tafjord, Oyvind},
  journal={arXiv preprint arXiv:1803.05457},
  year={2018}
}
